\documentclass[11pt,a4paper]{article}
\usepackage{authblk}
\usepackage[hyperref]{emnlp2020}
\usepackage{times}
\usepackage{latexsym}

\usepackage{microtype}

\newcommand{\bluetext}[1]{{\color{blue} \textit{#1}}}
\newcommand{\redtext}[1]{{\color{red} \textit{#1}}}

\newcommand{\egbox}[1]{
\smallskip
\noindent
\fbox{
        \parbox{0.95\linewidth}{
        \vspace{0.5ex}#1\vspace{0.5ex}}
      }
\smallskip
}

\aclfinalcopy 
\def\aclpaperid{77} 

\usepackage{times}
\usepackage{url}
\usepackage{latexsym}
\usepackage{graphicx}
\usepackage{amsmath}

\title{On Data Augmentation for Extreme Multi-label Classification}

\author[1]{Danqing Zhang}
\author[2]{Tao Li}
\author[1]{Haiyang Zhang}
\author[1]{Bing Yin}
\affil[1]{Amazon.com}
\affil[2]{University of Utah}
\affil[ ]{ \textit{\{danqinz,hhaiz,alexbyin\}@amazon.com,tli@cs.utah.edu}}

\date{}

\begin{document}
\maketitle
\begin{abstract}
In this paper, we focus on data augmentation for the extreme multi-label classification (XMC) problem.
One of the most challenging issues of XMC is the \emph{long tail} label distribution where even strong models suffer from insufficient supervision.
To mitigate such label bias, we propose a simple and effective augmentation framework and a new state-of-the-art classifier.
Our augmentation framework takes advantage of the pre-trained GPT-2 model~\cite{radford2019language} to generate label-invariant perturbations of the input texts to augment the existing training data.
As a result, it present substantial improvements over baseline models.
Our contributions are two-factored: (1) we introduce a new state-of-the-art classifier that uses label attention with RoBERTa~\cite{liu2019roberta} and combine it with our augmentation framework for further improvement;
(2) we present a broad study on how effective are different augmentation methods in the XMC task.

\end{abstract}

\section{Introduction} \label{sec:intro}

The extreme multi-label classification (XMC) problem aims to assign a set of inter-dependent labels to an input text. Such labeling can come in handy for industrial applications including product searching \cite{mcauley2015inferring}, document categorization \cite{mcauley2013hidden}, and social media recommendation \cite{jain2016extreme}. The \emph{extreme}ness is two-factored. In the task definition, it lies in the size of output space which ranges from few hundreds \cite{katakis2008multilabel,tsoumakas2008effective} to many thousands \cite{mencia2008efficient,zubiaga2012enhancing,partalas2015lshtc}. In practice, real-world data often has labels that are \emph{extremely} unbalanced. For instance, the AmazonCat-13K dataset is dominated by examples with media-related labels cover $\sim60\%$ of the data.

\begin{figure}[ht!]
    \centering
    \egbox{\bluetext{Original:} 1/24 scale the road warrior Mad Max Interceptor model kit...\\
    \bluetext{Generated:} 1/35 die-cast german panzerkamp- fwagen 1/35th scale die-cast model kit...\\
    \redtext{Labels:} \{play vehicles, toys \& games, vehicles \& remote-control\}} \vspace{0.5ex}
    \egbox{\bluetext{Original:}Aluminium case for cingular Nokia e62 / e61 (silver) with screen protector /SEP/ aluminium hard case is specially designed for Nokia e62 / e61 note:device is not included\\
    \bluetext{Generated:} samsung galaxy s6 edge /sep/ samsung galaxy s6 edge with screen protector ...\\
    \redtext{Labels:} \{accessories, cases, cases \& covers, cell phones \& accessories\}}
    \caption{Two examples of label-invariant perturbations generated by fine-tuned GPT-2 model. The generated texts that are conditioned on the original texts are kept in lower case.}
    \label{fig:intro:example}
\end{figure}

Recent works have shown impressive improvements in terms of accuracies with the emerging neural architectures~\cite[eg][]{devlin-etal-2019-bert,liu2019roberta}. However, they nonetheless are powered by large amount of training data and taking the unbalanced data as-is. As a result, even state-of-the-art models perform poorly on the tail labels (i.e. tail distribution) \cite{chang2019x}. This problem can be much more prominent when training data is limited, such as non-English corpus, or limited budget for data annotation.

One immediate approach to address the problem is data augmentation which can compensate the scarce data for tail labels.
To this end, we study two general augmentation strategies: 1) off-the-shelf rule-based augmentation, and 2) language-model-based augmentation.
Here we aim to compare how they impact the downstream XMC task.

For rule-based system, we examinate the EDA~\cite{wei2019eda} system and a simplified WordNet-only system.
For the language-model-based data augmentation, we propose a simple and effective approach called GDA (GPT-2 based data augmentation) to generate label-invariant examples using GPT-2 model. Specifically, given an unbalanced dataset, we first group examples pairs with the same label sets, then fine-tune the pre-trained GPT-2 to generate label-invariant perturbations. Such perturbed examples, such as shown in Fig~\ref{fig:intro:example}, are then used to augment the existing training data, particularly for those with tail labels.

In our experiments, we focus on the AmazonCat-13K dataset \cite{mcauley2013hidden} to validate the above augmentation approaches.
We sample different percentages of the training data to see how augmentation affects downstream XMC models from low data scenario to large data scenario.
We start with the vanilla RoBERTa~\cite{liu2019roberta} model and propose a new state-of-the-art model LA-RoBERTa by using label attention \cite{you2019attentionxml}.
We take both models as our base models and apply them on augmented datasets.
As a result, our augmented models outperform baseline (even the new state-of-the-art baseline) in terms of both overall precision and tail label precision.

In summary, our contributions include:
\begin{itemize}
    \item{We propose GDA, a simple and effective augmentation approach for the XMC task by using GPT-2 model. Our augmented system acchieves better performances on tail labels.}
    \item{We propose a new state-of-the-art classifier LA-RoBERTa, by combining RoBERTa and label attention it outperforms prior best model by $0.7$ in top-5 precision.}
    \item{We conduct so far the broadest study on data augmentation methods for the XMC task, showing that GPT-2 augmentation performs better when training data is rich, while rule-based system is relatively better when data is limited.}
\end{itemize}
\section{Model}
In this section, we will introduce the data augmentation framework for the XMC task. 
In Section~\ref{sec:generation}, we will present three augmentation models, two for rule-based and one for language-model-based. In Section~\ref{sec:classification}, we will present our state-of-the-art classifier. In Section~\ref{sec:training_with_aug}, we will cover how to use the generated examples for augmentation.

\begin{figure*}[ht]
  \includegraphics[width=1\textwidth]{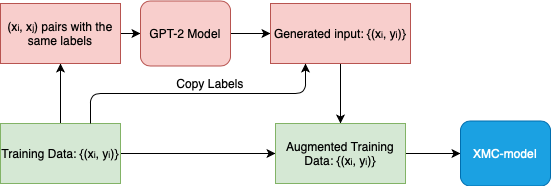}
  \caption{GPT-2 based data augmentation (GDA) framework}
  \label{fig:GDA-framework}
\end{figure*}

\subsection{Generation} \label{sec:generation}
Here, we study two types of augmentations: 1) rule-based; 2) language-model-based.
As noted in Section~\ref{sec:intro}, the training data of XMC is often very unbalanced. 
Here we focus on generating examples for tail labels.

\subsubsection{Augmenting with rule-based system}
We will briefly present two rule-base augmentation systems: EDA~\cite{wei2019eda} and a WordNet-based one.
Both systems utilize external knowledge base WordNet \cite{chris1998wordnet}, and take input from the given training data $D$ and make further editions. 

\paragraph{EDA}
EDA is a light-weight rule-based system that generates perturbations by random replacement, insertion, deletion, and swapping. The replacement operation uses WordNet \cite{chris1998wordnet} while others modify in-place. Examples generated in such way are often no longer natural, and should hypothetically benefit downstream models in terms of robustness.
Interestingly, prior works have shown that the generated examples can improve accuracy in multi-class classification task~\cite{wei2019eda}. In this paper, we aim to study its impact on the XMC task.

\paragraph{WordNet only}
To generate natural perturbations while keep the design simple, we simplify the EDA system by removing those random editing operations and only keep the synonym-swapping part. N non-stop words from the sentence are randomly chosen and are replaced with one of its synonyms chosen at random.



\subsubsection{Augmenting with pre-trained language model}

We use the GPT-2 model to demonstrate the use of language model for data augmentation. Fig~\ref{fig:GDA-framework} shows the overview of the language-model-based data augmentation framework.
\cite{kumar2020data} have shown successful application of label-to-text generation for multi-class classification task. However, such method does not directly apply to the XMC task due to the fact that labels in XMC are not order-dependent.
\cite{yang2020g} have used back-translation to augment existing training data which yields examples more like paraphrases.
However, as noted in Sec~\ref{sec:intro}, XMC examples that have the same label set can be substantially different.
Therefore, inspired by~\cite{anaby2020not}, we fine-tune a pre-trained GPT-2 model on paired XMC textual inputs.


Given a dataset $D = \{(x_i,y_i) : i=1,2,...,N\}$ and $K$ labels, where $x_i$ is the input sequence and $y_i \in R^{K}$ is the binary label vector of $x_i$, i.e. if $y_i^{k}=1$, then label k is one of the assigned labels of $x_i$. We group a set of example pairs $(x_i, x_j)$ that have the same assigned labels, i.e. $y_i=y_j$. We then fine-tune a pre-trained GPT-2 model to generate $x_j$ given $x_i$, where $(x_i, x_j)$ are subsamples from the training set of XMC (e.g. in Fig~\ref{fig:gpt_train_pair}). For training, we use the standard token-wise cross entropy loss with teacher forcing.
\begin{figure}[ht!]
    \centering
    \egbox{\bluetext{$x_i$:} Tom's of Maine Natural Moisturizing Body Wash 11.15 fl oz (330 ml) /SEP/  The product is not eligible for priority shipping\\
    \bluetext{$x_j$:}  Roger \& Gallet Jean Marie Farina Bath \& Shower Gel - 8.4 fl. oz.  /SEP/  Made in France.}
    \caption{An example $(x_i,x_j)$ for fine-tuning GPT-2.}
    \label{fig:gpt_train_pair}
\end{figure}

At inference time, we use the fine-tuned GPT-2 model to generate label-invariant perturbations for the rest training sets. Note that we omit the dominant label and only focus on generation for tail labels.

\subsection{Classification} \label{sec:classification}
To validate the impact of our augmentation framework, we will start with two base models: vanilla RoBERTa as a strong baseline and a new state-of-the-art model based on RoBERTa with label attention (LA-RoBERTa).
Let us first denote the augmented dataset $\widetilde{D}$.

\paragraph{RoBERTa} Given an example $(x_i, y_i) \in \widetilde{D}$ (in the absence of augmentation, $\widetilde{D}=D$), we score association between the input sequence $x_i$ and a particular label $y_i^k$ by:
\begin{align*}
    e_i &= \text{cls}(\text{RoBERTa}(x_i)) \\
    s_i^k &= f_k(e_i) \\
    \hat{y_i^k} &= \sigma(s_i^{k})
\end{align*}
where the $\text{cls}(\cdot)$ function returns the encoding of the [\emph{CLS}] token, $f_k$ is a label-wise linear layer without activation, $\hat{y_i^k}$ is the prediction probability for the $k$-th label.


\paragraph{Label Attention RoBERTa (LA-RoBERTa)}  We adopted the multi-label attention design from AttentionXML\cite{you2019attentionxml} and built it upon the vanilla RoBERTa. Specifically:
\begin{align*}
    \alpha_i^{tk} &= \frac{\exp^{{h_i^{t}} w_{k}}}{\sum_{l=1}^{T}  \exp^{{h_i^{l}} w_{k}}} \\
    m_i^{k} &= \sum_{t=1}^{T} \alpha_i^{tk} h_i^{t}\\
    s_i^k &= f_k(m_i^{k}) \\
    \hat{y_i^k} &= \sigma(s_i^{k})
\end{align*}
where $h_i^{t}$ is the encoding of the $t$-th token of input $x_i$, $\alpha_i^{tk}$ is the attention parameters for label $k$ on token $t$ for input $x_i$. Here we assume input $x_i$ consists of $T$ tokens including [\emph{CLS}] and [\emph{SEP}] tokens. By introducing label attention, each token embedding can have different impact on each label.


\subsection{Training with Augmented Data} \label{sec:training_with_aug}
We treat the original example and the generated ones differently, since in practice, there can be noise and minor errors in generated texts. 
Therefore, we use a weighted loss (parameterized by $\lambda$) to balance the augmentation data and the original one during training.
For examples from the existing data, we set $\lambda\texttt{=}1$ while grid-searched the best $\lambda$ for the generated example according to validation performances. Details are reported in Sec~\ref{sec:experiments}.



%

\section{Evaluation}
We use two sets of metrics to measure performances over \emph{all} labels and \emph{tail} labels.\footnote{For reference, evaluation scripts can be found at \url{https://github.com/kunaldahiya/pyxclib} and \url{http://manikvarma.org/downloads/XC/XMLRepository.html}.}

\paragraph{Metrics for \emph{all} labels}
We follow~\cite{Bhatia16} to use $P@k$ (Precision at k) and $N@k$ (nDCG@k: normalized Discounted Cumulative Gain at k) as our evaluation metrics. The two metrics are the most widely used evaluation metrics for XMC benchmarks. Specifically:
\begin{align*}
\text{P@k} &= \frac{1}{k} \sum^{k}_{i=1} y_{r(i)}\\
\text{DCG@k} &= \sum^{k}_{i=1} \frac{y_{r(i)}}{log(i+1)}\\
\text{iDCG@k} &= \sum^{min(k, ||y||_{0})}_{i=1} \frac{1}{log(i+1)}\\
\text{nDCG@k} &= \frac{\text{DCG@k}}{\text{iDCG@k}}
\end{align*}
where $r(i)$ denotes the rank of the $i$-th label in the top-k labels.
Note when $k\texttt{=}1$, we have $\text{P@1}\texttt{=}\text{DCG@1}\texttt{=}\text{nDCG@1}$. 

\paragraph{Metrics for \emph{tail} labels}
We follow~\cite{jain2016extreme} and use $\text{PSP@k}$ to describe the propensity precision of the top-k labels,
and $\text{PnDCG@k}$ to describe the normalized cumulative gain at top-k. Specifically,
\begin{align*}
\text{PSP@k} &= \frac{1}{k} \sum^{k}_{i=1} \frac{y_{r(i)}}{p_{r(i)}}\\
\text{PDCG@k} &= \sum^{k}_{i=1} \frac{y_{r(i)}}{ p_{r(i)}   log(i+1)}\\
\text{PSnDCG@k} &= \frac{\text{PDCG@k}}{\text{iDCG@k}}
\end{align*}
$p_{r(i)}$ is the propensity score for the rank of the $i$-th label, and the score is calculated from the training dataset.

Across our experiments in Sec~\ref{sec:experiments}, we will consider P@5 and PSP@5 the primary precision metrics, similarly, nDCG@5 and PSnDCG@5 the primary ranking metrics.


\begin{table*}[t]\centering
\scalebox{0.9}{
\begin{tabular}{ |c|c|c|c|c|c|c|c| } 
 \hline
Model & $\text{P@1}$ & $\text{P@3}$ & $\text{P@5}$ & $\text{nDCG@3}$ & $\text{nDCG@5}$ \\ 
\hline
XML-CNN  & 93.26 &	77.06 &	61.40 & 86.20 & 83.43\\
AttentionXML & 95.65 &	81.93 &	66.90 & 90.71 & 89.01\\ 
PfastreXML & 91.75 & 77.97 & 63.68 & 72.00 & 64.54\\
\hline
RoBERTa&  95.58 & 80.76 & 63.70 & 89.88 & 86.49\\ 
LA-RoBERTa & \textbf{96.28} &	\textbf{83.06} & \textbf{67.64} & \textbf{91.85} & \textbf{90.00}\\ 
 \hline
\end{tabular}
}
\caption{Model comparison on AmazonCat-13K using P@k and nDCG@k metrics.} \label{table:sota1}
\end{table*}

\begin{table*}[t]\centering
\scalebox{0.9}{
\begin{tabular}{ |c|c|c|c|c|c| } 
 \hline
Model &  $\text{PSP@1}$ & $\text{PSP@3}$ & $\text{PSP@5}$ & $\text{PSnDCG@3}$& $\text{PSnDCG@5}$ \\ 
  \hline
 XML-CNN &  52.42 &	62.83 &	67.10 & - & -\\
 AttentionXML  &  53.52 &	68.73 &	\textbf{76.38} &  - & -\\ 
 PfastreXML & \textbf{69.52} & \textbf{73.22} & 75.48 & 72.21 & 73.67\\
 \hline
 RoBERTa &  49.05 & 60.81 & 64.81 & 58.79 & 62.95\\ 
LA-RoBERTa & 51.01 & 66.52 & 74.86 & 63.28 & 69.91\\ 
 \hline
\end{tabular}
}
\caption{Model comparison on AmazonCat-13K using PSP@k and PSnDCG@k metrics. Fields marked with ``-'' are not provided in the orginal works.} \label{table:sota2}
\end{table*}

\section{Experiments} \label{sec:experiments}

In this section, we will explain how we subsample the AmazonCat-13K dataset in Sec~\ref{sec:dataset}.
Then we will present details of our rule-based and neural-based generators in Sec~\ref{sec:generators}, and RoBERTa baselines in Sec~\ref{sec:cls_baselines}.
Across our experiments, we aim to \emph{study the effectiveness of different augmentation methods with respect to the sizes of training data}.

\subsection{Dataset} \label{sec:dataset}
We follow the standard approach in~\cite{chang2019x,you2019attentionxml} to split the train/development sets of the AmazonCat-13K dataset. We set aside 10\% of the training instances as the validation set for hyper-parameter searching.
This results in training data of $1,067,615$ examples, validation set of $118,624$, and testing set of $306,782$.
Furthermore, we subsample $1$\%, $5$\%, $50$\%, and $100$\% of the effective training data to study in low-training and high-training scenario.
Finally, we use both of the product title and description as input text. For efficiency, we limit the maximal input length to $500$ words.


\subsection{Generators} \label{sec:generators}

For rule-based augmentation, we use the textattack package \cite{Morris2020TextAttack} for the EDA and WordNet Data Augmentation experiments.

For neural-based augmentation, we use the small version of GPT-2 model released by~\cite{wolf2019huggingface}. Furthermore, we set a thread on the input text length that we only keep the texts with 5 to 200 words. We also filter out the input pairs with high similarity score
\footnote{For reference, we used the SequenceMatcher from the difflib python package to calculate similarity score. \url{https://docs.python.org/2/library/difflib.html}}
(over 0.95) so that the model can focus on generating diversified examples instead of just paraphrasing.
For training, we set the learning rate $0.0001$ and beam search with width 10. For decoding, we set the temperature as 0 to smooth decoding probabilities and set the penalty to repeated prediction to $1$. We grid-searched hyperparameters by measuring the BLEU scores~\cite{papineni2002bleu} of fine-tuned model. In practice, we found such setting generally produces good examples.

As noted in Sec~\ref{sec:generation}, we focus on generating examples for tail labels. To this end, since media-related labels cover $\sim60\%$ of the data, we extract  media-related labels in Table~\ref{tab:label_table} and filter them out.
Furthermore, we found examples with media labels pose challenge to augmentation models since such examples are extremely diverse and often not even semantically related.

\begin{table}[ht!]
 \centering
\begin{tabular}{ |c|c| }
\hline
Label & Coverage \\ 
\hline
books & 24.88\%\\
music & 13.62\%\\
movies & 6.24\%\\
 tv & 8.99\%\\
games & 1.16\%\\
\hline
\end{tabular}
\caption{Media labels in the AmazonCat-13K dataset. We filtered out examples having these labels and focus on augmentation for tail labels} \label{tab:label_table}
\end{table}


\begin{table*}[t]\centering
\scalebox{0.85}{
\begin{tabular}{ |c|c|c|c|c|c|c|c|c|c| } 
 \hline
DA & $\%$Train & $N_{train}$ & weight&$\text{P@1}$ & $\text{P@3}$ & $\text{P@5}$ & $\text{nDCG@3}$ & $\text{nDCG@5}$ \\ 
  \hline
- & 100\% & 1,067,615 &  NA &95.58 & 80.76 & 63.70 & 89.88 & 86.49\\ 
GDA &100\%  & 1,284,855 & 0.5 & \textbf{95.71} & \textbf{81.39} & \textbf{64.70} & \textbf{90.38} & \textbf{87.35} \\ 
EDA & 100\% & 1,284,855 & 0.5 & 95.45 & 80.42 & 63.14 & 89.55 & 85.95\\
WordNet & 100\% & 1,284,855 & 0.5 & 95.48 & 80.34 & 63.04 & 89.47 & 85.86\\
\hline
- & 50\%& 533,807 & NA & \textbf{95.26} & 80.08 & 62.87 & 89.28 & 85.69\\
GDA & 50\% & 642,400 & 0.5 & \textbf{95.26} & \textbf{80.46} & \textbf{63.65} & \textbf{89.56} & \textbf{86.34} \\
EDA & 50\% & 642,400 & 0.5 & 95.21 & 80.00 & 62.79 & 89.16 & 85.55\\
WordNet & 50\% & 642,400 & 0.5 & 95.21 & 79.81 & 62.44 & 89.00 & 85.25\\
\hline
- & 5\%& 53,380 & NA & 91.61 & 74.63 & 58.01 & 84.02 & 80.23\\
GDA & 5\% & 64,297 & 0.5 & \textbf{92.91} & 76.24 & 59.27 & 85.59 & 81.73 \\
EDA & 5\% & 64,297 & 0.5 & 92.85 & \textbf{76.56} & \textbf{59.47} & \textbf{85.84} & \textbf{81.94}\\
WordNet & 5\% & 64,297 & 0.5 & 92.78 & 76.30 & 59.24 & 85.60 & 81.67\\
\hline
- & 1\% & 10,676 & NA & 89.68 & 71.09 & 54.04  & 80.79  & 76.23 \\ 
GDA & 1\% & 12,832 & 0.5 & 89.76 &71.50 &54.46 & 81.08 & 76.61\\
EDA & 1\% & 12,832 & 0.5 & \textbf{89.99} & \textbf{71.92} & \textbf{54.80} & \textbf{81.49} & \textbf{77.00}\\
WordNet & 1\% & 12,832 & 0.5 & 89.83 & 71.75 & 54.74 & 81.30 & 76.89\\
\hline
\end{tabular}
}
\caption{Performances on AmazonCat-13k \emph{all} labels of augmentation over RoBERTa baseline.} \label{roberta_1}
\end{table*}

\begin{table*}[t]\centering
\scalebox{0.85}{
\begin{tabular}{ |c|c|c|c|c|c|c|c|c|c| } 
 \hline
DA & $\%$Train & $\text{PSP@1}$ & $\text{PSP@3}$ & $\text{PSP@5}$ & $\text{PSnDCG@3}$ & $\text{PSnDCG@5}$ \\ 
  \hline
- & 100\% & 49.05 & 60.81 & 64.81 & 58.79 & 62.95\\ 
GDA & 100\% & \textbf{49.41} & \textbf{61.91} & \textbf{67.00} & \textbf{59.63} & \textbf{64.42} \\ 
EDA & 100\% & 49.22 & 60.43 & 63.75 & 58.57 & 62.31\\
WordNet & 100\% & 49.18 & 60.43 & 63.63 & 58.54 & 62.24\\
\hline
- & 50\%& 48.52 & 59.90 & 63.32 & 58.03 & 61.86\\
GDA & 50\%  & 48.77 & \textbf{60.69} & \textbf{65.24} & \textbf{58.62} & \textbf{63.10} \\
EDA & 50\% & 49.02 & 59.80 & 62.72 & 58.07 & 61.59 \\
WordNet & 50\% & \textbf{49.26} & 60.23 & 63.48 & 58.44 & 62.14 \\
\hline
- & 5\%& 46.67 & 55.39 & 57.94 & 54.28 & 57.53\\
GDA & 5\% & 47.67 & 56.82 & 59.14 & 55.56 & 58.71\\
EDA & 5\% & \textbf{47.69} & \textbf{57.11} & \textbf{59.45} & \textbf{55.77} & \textbf{58.94} \\
WordNet & 5\% & 47.59 & 56.85 & 59.13 & 55.56 & 58.67\\
\hline
- & 1\% & 45.58 & 52.00 & 52.36 & 51.64  & 53.57\\ 
GDA & 1\% & 45.80 & 52.32 & 52.95 & 51.90 & 53.99\\\
EDA & 1\% & \textbf{45.86} & \textbf{52.77} & 53.42 & \textbf{52.23} & \textbf{54.36}\\
WordNet & 1\% & 45.81 & 52.71 & \textbf{53.44} & 52.16 & 54.33\\
\hline
\end{tabular}
}
\caption{Performances on AmazonCat-13k \emph{tail} labels of augmentation over RoBERTa baseline.} \label{roberta_2}
\end{table*}

\subsection{Base Classifiers} \label{sec:cls_baselines}
Our baseline classifiers use the large version of pre-trained RoBERTa. We further fine-tune it on the AmazonCat-13K dataset and we manually tune the learning rate and the number of epochs. We train RoBERTa models using a learning rate of 1e-6, and 5e-6 for LA-RoBERTa models. For $100\%$, $50\%$, $5\%$ and $1\%$ of the data, we train the models for 10, 10, 40 and 100 epochs respectively.

In Table~\ref{table:sota1}, we present the performances of our baseline systems along with prior best models, i.e. XML-CNN \cite{liu2017deep}, AttentionXML\cite{you2019attentionxml}, and PfastreXML \cite{jain2016extreme}.
We see that the vanilla RoBERTa model is positioned strongly with cleaner model architecture.
With label attention, the LA-RoBERTa strongly outperforms the prior best model by $0.7$ in $\text{P@5}$ and $1.0$ in $\text{nDCG@5}$.

To our surprise, the AttentionXML achieves better $\text{PSP@5}$ than LA-RoBERTa while the later has much better $\text{P@5}$.
Note that the only major difference between AttentionXML and our LA-RoBERTa is BiLSTM encoder v.s. transformer encoder.\footnote{According to~\cite{you2019attentionxml}, this difference only applies to the AmazonCat-13K.}
In our preliminary experiments, we indeed found that BiLSTM encoder tends to have better $\text{PSP@k}$ performances than transformer encoder. But this comes at the cost of the $\text{P@k}$ metric.

In Table~\ref{table:sota2}, we show their precision scores for tail labels.
The new state-of-the-art LA-RoBERTa has subpar performances on tail labels compared to the PfastreXML, since PfastreXML optimizes over the propensity scored losses to get optimal peformance for tail labels.

\begin{table*}[t]\centering
\scalebox{0.85}{
\begin{tabular}{ |c|c|c|c|c|c|c|c|c|c| } 
 \hline
DA & $\%$Train & $N_{train}$ & weight&$\text{P@1}$ & $\text{P@3}$ & $\text{P@5}$ & $\text{nDCG@3}$ & $\text{nDCG@5}$ \\ 
\hline
- & 100\% & 1,067,615 & NA & 96.28 & \textbf{83.06} & 67.64 & \textbf{91.85} & 90.00\\ 
GDA & 100\% & 1,284,855& 0.5 & \textbf{96.29} & \textbf{83.06} & \textbf{67.69} & 91.84 & \textbf{90.03} \\ 
EDA & 100\% & 1,284,855& 0.5 & 96.19 & 82.85 & 67.41 & 91.64 & 89.74\\
WordNet & 100\% & 1,284,855& 0.5 & 95.95 & 82.58 & 67.11 & 91.32 & 89.39\\
\hline
-  & 50\% & 533,807 & NA & 95.73 & 82.12 & 66.66 & 90.94 & 88.95\\ 
GDA & 50\% & 642,400 & 1 & \textbf{95.79} & \textbf{82.25} & \textbf{66.78} & 91.06 & \textbf{89.07} \\
EDA & 50\% & 642,400 & 0.5 & 95.75 & \textbf{82.25} & 66.75 & \textbf{91.07} & \textbf{89.06}\\
WordNet & 50\% &  642,400 & 0.5 & 95.76 & 82.22 & 66.73 & 91.03 & 89.05\\
\hline
- & 5\% & 53,380 & NA & 92.36 & 75.05 & 58.6 & 84.52 & 80.90\\ 
GDA & 5\% & 64,297 & 0.5 &  92.40 & 76.74 & \textbf{60.61} & 85.82 & 82.68\\
EDA & 5\% & 64,297 & 0.5 & \textbf{92.78} & \textbf{76.81} & 60.52 & \textbf{85.97} & \textbf{82.71}\\
WordNet & 5\% & 64,297 & 0.5 & 92.63 & 76.62 & 60.40 & 85.78 & 82.56\\
\hline
- & 1\% & 10,676 & NA & 90.01 &	70.05 &	53.09 & 79.98 & 75.39\\ 
GDA & 1\% & 12,832 & 0.5 & \textbf{90.06} &	\textbf{70.90} & \textbf{54.32} & \textbf{80.69} & \textbf{76.52}\\
EDA & 1\% & 12,832 & 0.5 & 89.97 & 70.63 & 54.02 & 80.41 & 76.18\\
WordNet & 1\% & 12,832 & 0.5 & 89.71 & 70.36 & 53.87 & 80.12 & 75.95\\
\hline
\end{tabular}
}
\caption{Performances on AmazonCat-13k \emph{all} labels of augmentation over LA-RoBERTa baseline.} \label{awr_1}
\end{table*}

\begin{table*}[t]\centering
\scalebox{0.85}{
\begin{tabular}{ |c|c|c|c|c|c|c|c| } 
 \hline
DA &  $\%$Train & $\text{PSP@1}$ & $\text{PSP@3}$ & $\text{PSP@5}$ & $\text{PSnDCG@3}$ & $\text{PSnDCG@5}$ \\ 
  \hline
- & 100 \% & 51.01 & 66.52 & 74.86 & 63.28 & 69.91\\ 
GDA & 100\% & 51.29 & \textbf{66.73} & \textbf{75.08} & \textbf{63.49} & \textbf{70.12}\\ 
EDA & 100\% &  51.24 & 66.34 & 74.31 & 63.20 & 69.61\\
WordNet & 100\% & \textbf{51.70} & 66.42 & 73.90 & 63.37 & 69.48 \\
\hline
- & 50\% & 50.98 & 65.59 & 73.13 & 62.60 & 68.76\\
GDA & 50\%  & 50.93 & 65.74 & \textbf{73.55} & 62.70 & 69.01\\
EDA & 50\% & \textbf{51.36} & \textbf{66.03} & 73.38 & \textbf{63.03} & \textbf{69.07}\\
WordNet & 50\% & 50.92 & 65.70 & 73.30 & 62.67 & 68.88\\
\hline
- & 5\% & 47.46 & 56.49 & 59.16 & 55.28 & 58.60\\
GDA & 5\%  & \textbf{48.67} & \textbf{59.04} & \textbf{62.89} & \textbf{57.34} & \textbf{61.33}\\
EDA & 5\% & 48.49 & 58.84 & 62.51 & 57.16 & 61.06\\
WordNet & 5\% & 48.62 & 58.88 & 62.49 & 57.22 & 61.09\\
\hline
- & 1\% & 45.79 & 52.46 & 53.01 & 51.99 & 54.05\\ 
GDA & 1\% & \textbf{46.28} & \textbf{52.73} &	\textbf{53.71} & \textbf{52.32} & \textbf{54.64}\\
EDA & 1\% & 46.25 & 52.56 & 53.41 & 52.17 & 54.41\\
WordNet & 1\% & 46.16 & 52.39 & 53.25 & 52.01 & 54.24\\
\hline
\end{tabular}
}
\caption{Performances on AmazonCat-13k \emph{tail} labels of augmentation over LA-RoBERTa baseline.} \label{awr_2}
\end{table*}

\subsection{Classifiers with Data Augmentation}
In Table~\ref{roberta_1} and~\ref{roberta_2}, we show the impact of different augmentation methods on the vanilla RoBERTa baseline.
We see that our proposed GDA method is more effective than rule-based ones when training data is very large.
For instance, with $\sim$1M training examples, GDA outperforms the baseline by $1.0$ in $\text{P@5}$ and $0.9$ in $\text{nDCG@5}$.
For tail labels, GDA also improve over baseline by $3.2$ in $\text{PSP@5}$ and $1.5$ in $\text{PSnDCG@5}$.

To our surprise, rule-based methods failed to improve over baseline with large training data.
We hypothesize this is because rule-based augmentations could not bring much richer vocabulary diversity and, often, they yield grammatically incorrect texts.
In contrast, our GDA framework can generate substantially diversified input while maintaining natural inputs in general.\footnote{Generation is an evolving field. At this point, even strong models sometimes still output artificial texts, e.g. repeated phrases and off-context phrases.}

When training data is limited such as $1$\% of training data ($\sim$10k examples),
both rule-based augmentation and our GDA yield outperform baseline in $\text{P@k}$ and $\text{N@k}$ metrics.
And rule-based models yield more improvement than GDA.
We should note that such small amount of data brings challenge to fine-tune GPT-2 model, especially considering the label space is $13$k and the label distribution is extremely biased.
As a result, rule-generated examples, while being unnatural often, outperform those generated by GPT-2 models.

For the much stronger baseline LA-RoBERTa, we present similar comparison in Table~\ref{awr_1} and~\ref{awr_2}.
In general, we see similar observations as above when comparing different augmentation methods with respect to training sizes.
Note that, when using $100$\% of the training data, GDA only performs on par with the baseline when measuring the overall label precisions (i.e. $\text{P@k}$ and $\text{nDCG@k}$) while performances on tail labels are still better (e.g. $0.2$ improvement in $\text{PSP@5}$ and $0.1$ in $\text{PSnDCG@5}$).
Again, rule-based systems failed to improve precisions for both \emph{all} and \emph{tail} labels.

\paragraph{When and which data augmentation is effective in the XMC task?}
Given the above observations, we conclude that, when training data is very limited, both rule-based augmentation and GDA work better than base models.
When training data is rich, DGA still improves over baseline while rule-based systems start to hurt precisions.
Therefore, we recommend GDA since it improves more consistently against different training sizes.

\section{Related Works}

\subsection{Extreme Multi-label Classification}
Many prior XMC algorithms are based on sparse linear models that use TF-IDF features and label partitioning methods to reduce label space complexity. 

Recent works have focused on applying neural models to encode textual semantics (either query or document).
This resulted in substantially better performances than using discretized features.
XML-CNN \cite{liu2017deep}, first deep learning paper in this area, used CNN and dynamic pooling to learn textual representation. AttentionXML \cite{you2019attentionxml} used BiLSTM and multi-label attention to capture better interaction between input tokens and individual output labels. X-BERT \cite{chang2019x} fine-tuned pre-trained transformer models for the XMC task.

Prior works also focused on making more accurate tail label prediction. PfastreXML \cite{jain2016extreme} optimizes over propensity scored objective function. Adversarial XMC \cite{babbar2018adversarial} proposes to use regularized optimization objective.

\subsection{Data Augmentation in NLP}
Data augmentation has shown successful application in the computer vision field, and have been a rising field in natural language processing.
Common data augmentation techniques include rule-based data augmentation (e.g. EDA~\cite{wei2019eda}) and seq2seq model (e.g. back translation \cite{wei2019eda,xie2019unsupervised,xia2019generalized}, RNN-based variational autoencoder\cite{bowman2016generating}, and models based on conditional language model~\cite{kumar2020data,anaby2020not,yang2020g}).

Recent works have successfully applied data augmentation to NLP tasks like relation extraction \cite{papanikolaou2020dare}, spoken language understanding \cite{peng2020data} and text classification \cite{wei2019eda,anaby2020not}. But to the best of our knowledge, the impact of data augmentation has not been studied for the XMC task.


\section{Conclusion \& Future Work}
In this paper, we conduct so far the broadest study on data augmentation methods for the XMC task. The results are promising and demonstrate that both the strong baseline vanilla RoBERTa and state-of-the-art LA-RoBERTa can benefit from various data augmentation methods. Our proposed GDA improves both the precision and ranking metrics, and have even larger improvement for tail labels. We also observe rule-based synonym replacement data augmentation method demonstrates good performances when training data is scarce.
One potential next step is to further differentiate erroneous generations from good ones.
Another possibility is to dynamically assign the loss weight $\lambda$ at instance level for the generated data.

\bibliographystyle{acl.bib}
\bibliography{citations}

\end{document}